\documentclass[a4paper]{article}

\usepackage{ttr}
\usepackage{graphicx}
\usepackage{todonotes}
\usepackage[round,authoryear]{natbib}
\usepackage{hyperref}
\usepackage{xurl}

\title{Replicating and Extending Weight-Space Programming: Frontier LLMs Compile Programs into Weights and Read Input Layers}
\author{Roksana Goworek\thanks{All authors equally contributed.} \and Alexander Hickey \and Federico Nanni \and Martin O'Reilly}

\date{The Alan Turing Institute}

\begin{document}

\setcounter{page}{1}

\maketitle

\begin{abstract}
Researchers at MIT recently reported that GPT-6 Astra, a frontier large language model (LLM), can write the weights of small neural networks directly in a single reply, with no tools, and can read the raw weights of small networks to say what they were trained to do. In a three day sprint we replicated and extended these results, and we present our preliminary results and evaluation in this working note. We reached four main findings.
(1)~\emph{Writing weights reproduces.} Astra's networks for the game Breakout scored 514, 550 and 307 across three runs (the original reported 517), and scored just as well on two modified versions of Breakout that exist nowhere else, so it is not recalling networks from its training data. Claude Opus 5.5 can also write working networks, but often runs out of output tokens before it does.
(2)~\emph{What is new is translating programs into weights.} Earlier models could write a good policy as Python code roughly one generation before they could write it as network weights, and the networks the models write behave like compiled programs: two models independently decompiled Astra's Breakout network into Python that chooses the same action as the network in all 49,189 game states we tested. On a task with no program to compile, classifying handwritten digits, Astra's networks scored 50\% on average, below a simple template classifier fitted on the data (82\%).
(3)~\emph{Reading weights is mostly reading the input layer.} Astra and Opus 5.5 identify which game a network plays almost perfectly, and fall to chance when the weights are shuffled. But on three new games we designed and trained ourselves, Astra recovered where things are on the board (information a simple heatmap also shows to a human) while mostly missing how the game works.
(4)~\emph{Auditing from weights alone is possible but shallow.} Reading plain-text weights without running any code, Astra picked out a small language model poisoned during training in 10 of 10 runs, one poisoned by extra training afterwards in 7 of 10, and a malicious adapter in 23 of 32 cases, but it never found the trigger or described the malicious behaviour. Open-weight models lag behind, although Qwen3.8-27B succeeded on the easiest auditing task. From this early study, we conclude that LLMs can genuinely read and write small networks, but that this currently amounts to compiling hand-designed programs and reading the layout of the input layer. It is a promising but early tool for interpretability.
\end{abstract}

\section{Introduction}
Recently, researchers at MIT's Computer Science and Artificial Intelligence Laboratory (CSAIL) \citep{takehi2026weightspace} posted some intriguing initial results showing that OpenAI's latest frontier large language model (LLM), GPT-6 Astra, can write a 20,486 weight neural network that plays a version of the 1980's computer game Breakout, scoring 517 against 25 for a network of the same architecture trained for 500 million steps with Proximal Policy Optimisation (PPO), a standard reinforcement learning algorithm.\footnote{The PPO scores are those reported by the original authors (best of
several seeds); their networks were not released and we did not retrain them.
These baselines are probably weak: tuned PPO reaches far higher scores on
MinAtar Breakout. For example, \citet{kong2024generalizable} report a return of
679.3 $\pm$ 15.0 at the end of training with a fine-tuned PPO. Their setup and
the way they compute scores may differ from the original post's, so the
numbers are not directly comparable. Throughout this note, `PPO' refers to the original authors' baseline.\label{fn:ppo}} Astra wrote the network weights directly in a single reply, with no use of tools - that means no way to write or run code, no way to train the network in the usual manner, and no way to run versions of the network to inspect outputs and iterate. As far as we know, it is the first time a general-purpose model, used off the shelf, has written a working network like this from scratch. In the same post, the researchers reported that Astra can also read the raw weights of small networks and say what each was trained to do, edit a small language model's weights to remove a fault, and write a network that plays Go with a budget of at most 50 forward passes. Given the attention these claims have received, we examined the work more closely during a three-day sprint to separate the (understandable) online hype from what is genuinely new and relevant to our field. This working note presents our preliminary findings: what reproduces (Section~\ref{sec:exciting}), what we would temper (Section~\ref{sec:temper}), what weight reading could offer for auditing models (Section~\ref{sec:audit}), and a few research directions we think are worth pursuing (Section~\ref{sec:future}).

\section{What's exciting}
\label{sec:exciting}

We reproduced the original weight-writing and weight-reading experiments, compared earlier models to see when the ability appeared, and examined how the models write weights.\smallskip

\noindent\textbf{Key findings.}
\begin{itemize}
  \item \textbf{Writing weights reproduces.} Astra's Breakout networks in our replication experiments scored 514, 550 and 307, in line with the 517 score reported in the MIT results, and it scored just as well on two modified versions of Breakout that exist nowhere else. Claude Opus 5.5 can do it too, but often runs out of output tokens first.
  \item \textbf{Reading weights reproduces.} Astra and Opus 5.5 named the game for 12 of 12 networks, and fell to chance when the weights were shuffled.
  \item \textbf{What is new is translating programs into weights.} Earlier models could write a strong policy as Python code a generation before they could write it as weights (Table~\ref{tab:programs-vs-networks}).
  \item \textbf{The written networks are compiled programs.} Two models independently decompiled Astra's Breakout network into Python that matches its action in all 49,189 game states tested.
  \item \textbf{Open-weight models are not there yet}, at either writing or reading weights.
  \item \textbf{The ability probably comes from post-training}, not from new base models, and appeared in steps between late 2025 and September 2026.
\end{itemize}

We replicated the weights-writing experiment, which asks GPT-6 Astra for networks that can play three games from MinAtar, a miniature version of the Atari benchmark: Breakout, Seaquest and Asterix. Across three runs on Breakout, Astra's networks scored 514, 550 and 307, in line with the 517 score reported in the MIT results. On Asterix it averaged 66, a little higher than the 53 reported in the MIT results. On Seaquest it reached 26, about half the MIT result's single run of 53. Others have also reproduced these results (\cite{krentsel2026, dang2026}). 
To check that Astra was not recalling a Breakout network from its training data, we also asked the agents for two versions of Breakout that exist nowhere else: one with left and right swapped, and one with the board upside down. The network Astra wrote for the standard Breakout game scores just 0.5 and 0.0 on these variations, while the networks it wrote for these custom variations averaged 515 on both, the same as on the standard game, suggesting it is working from the game's code. Claude Opus 5.5, Anthropic's current frontier publicly available model, can do it too: two of its three networks for the swapped game scored about 450. But at the post's reasoning effort it often spends its whole output budget thinking and never writes the weights (5 of 9 runs).

The reading weights experiments reproduce as well. Shown the weights of the post's four game-playing networks, Astra and Opus 5.5 named the game every time (12 of 12 over three runs each). Shown three small diffusion models, they named the shape each one draws in 9 of 9 and 8 of 9 cases. With the weights shuffled within each matrix, both fell to chance, so they are reading the weights, not guessing from what they know about the games.

We then evaluated earlier models from the same two labs. To distinguish a model that does not know how to play from one that cannot encode its strategy into weights, we asked each model for the same policy as a Python program. We used the same prompt, except that it asks for a program instead of weights. 
\begin{table}[t]
\centering
\caption{\textbf{Models could write a game-playing strategy as code before they could write it as network weights.} Each model was asked to write a policy for three MinAtar games (miniature versions of Atari games) in a single reply, with no tools: either as a Python program (Code) or directly as the weights of a neural network with 600 inputs (a 10$\times$10 grid with six object channels), two hidden layers of 32 units and 6 output actions (Weights). Scores are the mean over 1,000 plays of each game; higher is better. By July 2026, Claude Opus 5 and GPT-5.6 Sol wrote strong programs (e.g.\ 580 on Breakout, 301 on Asterix) but weak networks; GPT-6 Astra is the first whose networks keep half to two-thirds of its programs' scores, beating the original authors' PPO-trained networks on Breakout and Asterix (which are probably under-tuned; see footnote~\ref{fn:ppo}). Code: one run per game. Weights: mean of 3--5 runs (Opus 5 and GPT-5.2: one run). PPO: scores reported by the original authors for networks of the same architecture, trained with Proximal Policy Optimization, a standard reinforcement learning algorithm, for 100 or 500 million environment steps.}

\label{tab:programs-vs-networks}
\setlength{\tabcolsep}{4pt}
\renewcommand{\arraystretch}{1.15}
\resizebox{\linewidth}{!}{%
\begin{tabular}{|l|c|rr|rr|rr|}
\hline
 & & \multicolumn{2}{c|}{Breakout} & \multicolumn{2}{c|}{Seaquest} & \multicolumn{2}{c|}{Asterix} \\
\cline{3-8}
Model & Released & Code & Weights & Code & Weights & Code & Weights \\
\hline
Claude Opus 4.5     & Nov 2025 &   6.4 &   3.4 &   1.8 &  0.1 &  31.5 &  0.8 \\
GPT-5.2             & Dec 2025 & --$^{c}$ & 3.5 & --$^{c}$ & --$^{c}$ & 30.1 & --$^{c}$ \\
Claude Opus 5$^{a}$ & Jul 2026 & 580.2 &  30.9 &   7.9 &  0.4 &  25.1 & 18.5 \\
GPT-5.6 Sol         & Jul 2026 &  78.7 &  31.1 &  19.4 &  0.8 & 300.8 & 14.6 \\
Claude Opus 5.5     & Sep 2026 & 626.1 & 112.0$^{b}$ & 179.9 & 0.0$^{b}$ & 74.1 & 45.1 \\
GPT-6 Astra         & Sep 2026 & 695.7 & 456.9 & --$^{c}$ & 25.5 & 127.0 & 66.2 \\
\hline
PPO, 100M steps & -- & \multicolumn{2}{c|}{15.8} & \multicolumn{2}{c|}{5.6}  & \multicolumn{2}{c|}{25.8} \\
PPO, 500M steps & -- & \multicolumn{2}{c|}{25.3} & \multicolumn{2}{c|}{88.9} & \multicolumn{2}{c|}{55.2} \\
\hline
\end{tabular}%
}
\par\smallskip
\begin{minipage}{\linewidth}\footnotesize
\textsuperscript{a}At \texttt{high} reasoning effort; at \texttt{xhigh} it ran out of
output tokens before writing any weights in 10 of 12 runs.
\textsuperscript{b}Runs that ran out of output tokens before writing any weights count
as 0 (2 of 3 on Breakout, 3 of 3 on Seaquest); its one finished Breakout network scored 336.
\textsuperscript{c}Lost to an application programming interface (API) failure.
\end{minipage}
\end{table}

As can be seen in Table~\ref{tab:programs-vs-networks}, the late-2025 model we could test fully, Claude Opus 4.5, mostly could not even write good programs. Its contemporary GPT-5.2 could not be tested fully: the API dropped most of our requests after 40 minutes to 2 hours of reasoning. Its one program, for Asterix, scored 30.1, like Opus 4.5's 31.5, and its one network, for Breakout, scored 3.5.

By July 2026, Opus 5's Breakout program scores 580 and GPT-5.6 Sol's Asterix program 301, but their networks still score only about 31 and 15. Astra, released in September 2026, writes networks that keep half to two-thirds of its programs' scores. What's new is translating programs into weights.

Open-weight models lag behind the frontier. In our limited experiments, even the largest open-weight models struggled with writing good game-playing networks. Kimi-K3.1-Pro scores 19 to 68 on Breakout, competitive with the PPO baselines reported in the MIT posts but well below Astra's 457 and the 439 Kimi itself averages when asked for a game-playing policy as a Python programme. Its networks also scored poorly on Seaquest (1 to 2) and Asterix (9 to 42), below the scores for both games obtained via PPO trained for 500 million steps. DeepSeek-V4-Pro's networks score 2 to 12 on Breakout and close to zero on Seaquest and Asterix, below PPO trained for 100 million steps. Qwen3.8-27B, a much smaller model, averages below 5 points on Breakout, Seaquest, and Asterix. Its highest score across 96 runs on Breakout was 41, while it was able to average 129 by using a Python policy.

On reading weights, Kimi-K3.1-Pro correctly identifies the game networks 11 of 12 times, identifying Breakout, Freeway and Pong in all 3 repeated runs, and identifying Pacman in 2 of 3 runs. It misidentifies the remaining Pacman run as a game with similar structure (Sokoban, a maze-based game where the objective is to move boxes onto specified spots in the maze to get rewards). However, it does less well on identifying the shapes produced by the diffusion networks. It identifies the circle-generating network in all 3 runs but only identifies the triangle-generating network on 1 of 3 runs and fails to identify the square-generating network at all. The previous generation of Kimi (Kimi-K3), does much worse. It names the correct game 6 of 12 times (above chance and the 2/12 it scores on shuffled controls), but can only reliably name Freeway (3 of 3 runs). It is unreliable on the other games, naming Breakout, Pacman and Pong correctly only once out of three runs for each game (and for Pacman incorrectly claiming it was a maze-free variant). DeepSeek-V4-Pro names the game or shape in 11 of 28 cases, though it is likely answering from its priors as it does about as well on shuffled weight controls (4 of 14), and it answered `Pac-Man' 7 times out of 8 for the shuffled game network controls, at a median confidence of 80\%.

Others have reported the much smaller Qwen3.8-27B correctly identifying the game-playing networks~\citep{willison2026qwen}. However, in our experiments, this smaller model was not consistently correct - across 33 runs per game it named the correct game a little under 40\% of the time (50/132). That said, it performed better than both chance and the shuffled control (17/116). This suggests that it is on average reading some signal from the networks, rather than relying purely on its own priors. For Breakout and Pacman it picked the correct game significantly more often than its next favourite choice (15 vs 7 for Breakout, 14 vs 6 for Pacman), though for Freeway (12 vs 11) and Pong (9 vs 10) this was not the case.

\subsection{How are the models doing this?}
\label{subsec:how_code_weights}
In short, the models appear to design a rule-based strategy to play the game and then translate it into weights, producing sparse networks whose units act as on/off switches. Compiling programs into model weights is an active area of research, but so far it has only been done mechanically or by hand. Tracr~\citep{lindner2023tracr} compiles programs written in RASP (Restricted Access Sequence Processing, a programming language designed to describe what transformers compute) into transformer weights. Based on the design notes the frontier LLMs include in their responses prior to outputting the network weights, they informally adopt similar strategies (although Astra's reasoning is hidden). From a description of the game and its source code, they decide what each hidden unit should compute, then set the weights to compute it. In the post, Astra's Breakout network uses its first layer to work out where the ball will land relative to the paddle, and its second to move the paddle there. It implements this with surprisingly few weights: only 815 of the 20,486 are non-zero, and only 6 of the 32 first-layer units are used at all. The weights into the second layer are all in the hundreds, up to 1,000, so its units behave less like smooth functions and more like programmatic on/off switches. Writing networks this way gives us a clear explanation of each element, because the agent designs and describes the features before it sets the weights, unlike in traditional model training, where the exact functions of individual components are almost impossible to disentangle.

Networks with known mechanisms provide useful controlled tests for mechanistic interpretability techniques. That is why Tracr was built, as `a laboratory for interpretability'. Networks written by LLMs could make such test cases more varied and on more realistic tasks, although their relevance to normally trained networks would still need to be established. For now, code is still the better and more efficient way to express a strategy, and every model's programs score higher than the networks. Neural networks are good at solving fuzzy, hard-to-define tasks, can be trained further, or combined with other networks. However, the ability of LLMs to write such networks remains to be demonstrated.

\subsubsection{Decompiling network weights back into code}
We suspected that the models design a strategy and then compile it into weights, because the two networks that play well (Astra's and Opus 5.5's) set only 4\% and 6\% of their weights, with median magnitudes of 10 and 30 and largest values of 1,000 and 108, while a network trained with PPO sets nearly all of its weights, with a median of 0.08 and a largest value below 9.

We confirm that the agents adopt this strategy-to-weights method by reverse-engineering it.
We asked two agents to independently translate Astra's Breakout network into Python code. Claude Opus 5.5 and Astra each worked in a separate folder outside our repository, supplied with the network’s weights and architecture, the game’s rules, source code and input/output specification from the original prompt, and the decompilation task. They were told not to look elsewhere or use the web. They could analyse the weights with Python, but could not execute the network or any of its layers on inputs, or simulate the game. They therefore had to infer the policy from the weights and supplied specifications, without querying the network’s behaviour. Both ran at extra-high reasoning effort: Opus 5.5 finished in about six minutes and 37,000 output tokens, Astra in about four minutes and 9,000 tokens.

Both programs choose the same move as the written network in 100\% of the 49,189 game states we tested, and over 1,000 games they score 502.9 on average, exactly as the network does. The two models also read the network the same way, and their reading matches the design Astra described when it wrote it: two units track which way the ball is moving, four predict the column where it will land after bouncing off the walls, and the second layer moves the paddle towards that column. 
Neither found any connection to the bricks: the network ignores them. 
Even when told nothing about the game, Opus 5.5 and Astra decompiled the network into programs that reproduce its outputs in every board situation, inferring that the inputs form a 10×10 grid (Opus 5.5 even named the ball, its trail and the paddle), at about 46,000 and 21,000 output tokens. 
A network written this way is not a black box, and can be read back into the program behind it.

Programs decompiled by the same two models, from the Breakout networks of Opus 5.5, GPT-5.6 Sol and Opus 4.5 (scoring 336, 14 and 1), also matched them in every game state we tested, flaws included, which shows that these networks also encode an explicit plan rather than the distributed representations of a trained network. Astra's and Opus 5.5's networks follow similar plans but are wired differently. Astra's decision units, with weights of 200 to 1,000, always switch fully on or off, so its programs match it even on impossible board states, whereas Opus 5.5's, with weights of $\pm 3$, can be translated exactly into rules only on possible games.
And as a side benefit, we show that translating these sparse weights into code is at least as easy as writing them, since GPT-5.6 Sol, which cannot yet write a strong Breakout network itself, decompiled Astra's network, matching it in every game state.

\subsection{Where this ability might come from}
Weight-space abilities appeared in steps between late 2025 and September 2026. In our runs, the late-2025 models barely read weights and could not write them. Opus 4.5 mostly guessed, answering `Freeway' for 19 of 20 game networks. The July 2026 models read weights well (Opus 5 got 28 of 29 right) but still wrote weak networks. Astra, released in September 2026, wrote strong ones. Opus 5.5, released the same month, could too, though it often ran out of output tokens before writing any weights.

These abilities were first reported by MIT researchers rather than the model providers, suggesting that they might have emerged as a side effect of improving other capabilities rather than as an explicit training target. We also suspect that the gains came from post-training rather than new base models. We cannot see how the labs train their models, but tokenisers offer an external clue: a new base model often comes with a new tokeniser, while further training of the same model usually keeps the same one. We sent the same twelve short texts to each model and compared token counts. GPT-5.2, GPT-5.6 Sol and Astra counted them identically, as did Opus 4.8, Opus 5 and Opus 5.5. Both labs’ steps in writing weights, and OpenAI’s step in reading them, came without a new tokeniser. This is not proof, since a lab can pretrain a new model with the old vocabulary, but it fits the simpler explanation that the same base models were trained further.

Reinforcement learning on coding tasks is one likely candidate. It rewards agents for sticking with long tasks, breaking them into smaller parts and putting the pieces back together into something that works. That persistence may help with both the mechanistic analyses in the post and turning a planned behaviour into switches and gates in a network’s weights. That would fit what we saw: models wrote the strategy as code a generation before they could write it into weights. The models may also simply know more mechanistic interpretability than they used to. The field has grown quickly in recent years, so newer models have likely read far more of its papers and code, and may even have been post-trained on similar tasks.
The post hints at this: analysing small character-level language models from their weights alone, Astra found the heads that copy earlier text by multiplying weight matrices, which is, in the authors' words, `a method presented by Elhage et al., which we never mentioned to Astra'~\citep{takehi2026weightspace}.

It is exciting that a capability like this can appear without, as far as we can tell, anyone setting out to build it. There are few examples of it online, and few people have ever tried it by hand. But excitement is also a reason to be careful about what the experiments actually show.

\section{Where we would temper the claims}
\label{sec:temper}

The main results are reproducible, but both abilities are narrower than they first appear: writing works best when the task can be solved with programs to compile, and much of reading can be read off the first layer, by a model or by a person.\smallskip

\noindent\textbf{Key findings.}
\begin{itemize}
  \item \textbf{On a fuzzy task, written weights fall well short.} Astra's hand-written classifiers for handwritten digits scored 50\% on average, below a simple template classifier fitted on the data (82\%) and far below training (97\%).
  \item \textbf{Several claims are tested less strongly than their framing suggests}: most conditions were run once, `native' reading often involved Python or forward passes, and the more speculative claims were not tested.
  \item \textbf{Reading a game network is mostly reading its input layer}, which records where each kind of object appears on the board. A simple heatmap shows this to a human too.
  \item \textbf{On three new games we trained ourselves, Astra recovered the layout but mostly missed the mechanics}, such as which action does what, and often named a familiar game when the mechanics did not fit.
\end{itemize}

Our runs reproduce the main results, but not every claim in the original post is tested as strongly as its framing suggests, and several critiques have made this point within days of its release. Matthew Honnibal points out that `natively reading weights' covers very different setups: only three experiments work from raw weights alone, while every language-model experiment uses Python and, for Qwen, forward passes~\citep{honnibal2026astra}. The Qwen maths gain comes from stopping the model looping rather than improving its maths ability, and it shrinks sharply under other sampling settings~\citep{pyke2026weightspace}. Writing networks from scratch amounts to compiling a hand-designed program into weights, which is what our code-before-weights result suggests too~\citep{maier2026weightspace}. Most conditions were run once, and the more speculative claims (auditing models for backdoors, recursive self-improvement) are not tested at all.

\subsection{Writing weights works best when there is a program to compile}
\label{subsec:fuzzy}

The MinAtar games reward near-symbolic programs: the input is a grid of zeros and ones, and a few rules (track the ball, move under it) play well. Most tasks that networks are trained for have no such program. To test whether writing weights goes beyond compiling one, we asked Astra for a network that classifies handwritten digits (MNIST), under the same conditions as before: one reply, no tools, every weight starting at zero, and no access to the images. The network has the same two hidden layers of 32 units as the game networks, with 784 inputs (a 28$\times$28 image) and 10 outputs, and is scored by its accuracy on the 10,000 test images.

A network of this architecture trained on the 60,000 training images reaches 96.7\%. A much simpler baseline, which compares each image with the average image of each digit, reaches 82.0\%. Astra's networks scored 49.6\% on average over three runs (55.7, 47.2 and 46.0\%). It designed them the same way each time: a first layer of coarse ink detectors (regions, strokes, loop interiors, bars) and a second layer that matches them against several hand-drawn variants of each digit. They recognised 0s and 1s well (74--86\%) but 9s poorly (17--23\%).

To check that Astra was not recalling a network, we also asked for one for digits rotated a quarter turn, with the rotation stated in the prompt. A network written for upright digits scores 12\% on rotated ones. Astra's networks for rotated digits scored 49.4\% (57.5, 44.4 and 46.4\%), the same as for upright digits, and each scored only 18--22\% when given upright digits, so they were written for the rotated input. 

So Astra can turn vague knowledge of what digits look like into working weights, well above chance, but it falls short even of the simple baseline that sees the data, and far short of training. Its game networks beat the original post's reinforcement-learning baselines; on digits, a few lines of averaging beat it.

We then turn to the claim we think matters most: that frontier models can tell what a network was trained to do just by reading its weights.

\subsection{Reading weights is mostly reading the input layer}

In the original game experiments, every network takes the same 10$\times$10 grid, split into input channels labelled by role: agent, moving objects, reward targets, obstacles. Each weight in the first layer is therefore tied to a known grid cell and a known kind of object. That makes the first layer close to a picture of where each game cares about things.

\begin{figure}[!htb]
  \centering
  \includegraphics[width=\linewidth]{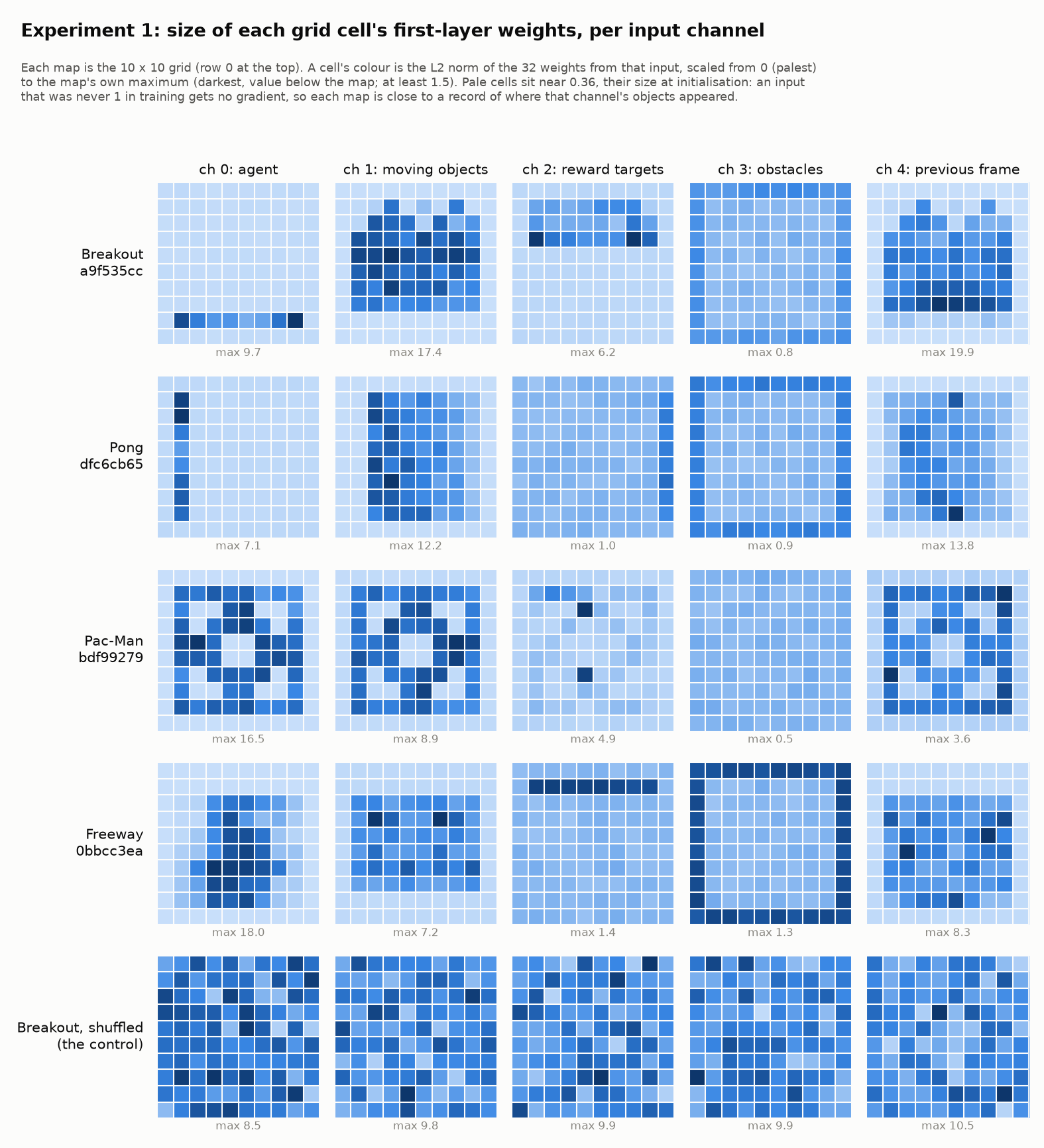}
  \caption{First-layer weight norms per input channel for the four original games. Each map is the 10$\times$10 grid; a cell's colour is the size of the 32 weights leaving that input, scaled independently from the minimum (palest) to the maximum (darkest) weight for each input channel for each game. The bottom row is the Breakout network with its weights shuffled, as in the original MIT control.}
  \label{fig:human-heatmaps}
\end{figure}

\paragraph{Humans can see it too.} Sean Goedecke showed that a few lines of NumPy computing per-cell weight norms turn the first layer into a heatmap~\citep{goedecke2026superhuman}. Figure~\ref{fig:human-heatmaps} shows the same for our networks. Once plotted, the paddle along the bottom row in Breakout or the maze in Pac-Man is easy to spot without any game label. The agent channel alone narrows things down: a row for Breakout's paddle, a column for Pong's, the corridors of a maze for Pac-Man, and a vertical band up the middle for Freeway's chicken, whose reward row sits along the top. The pattern comes from training: an input cell that was never active gets no gradient and keeps its small initial weights, so each map is close to a record of where that channel's objects appeared. Shuffling the weights, as the original control does, erases it (bottom row). 

\begin{figure}[ht]
  \centering
  \includegraphics[width=\linewidth]{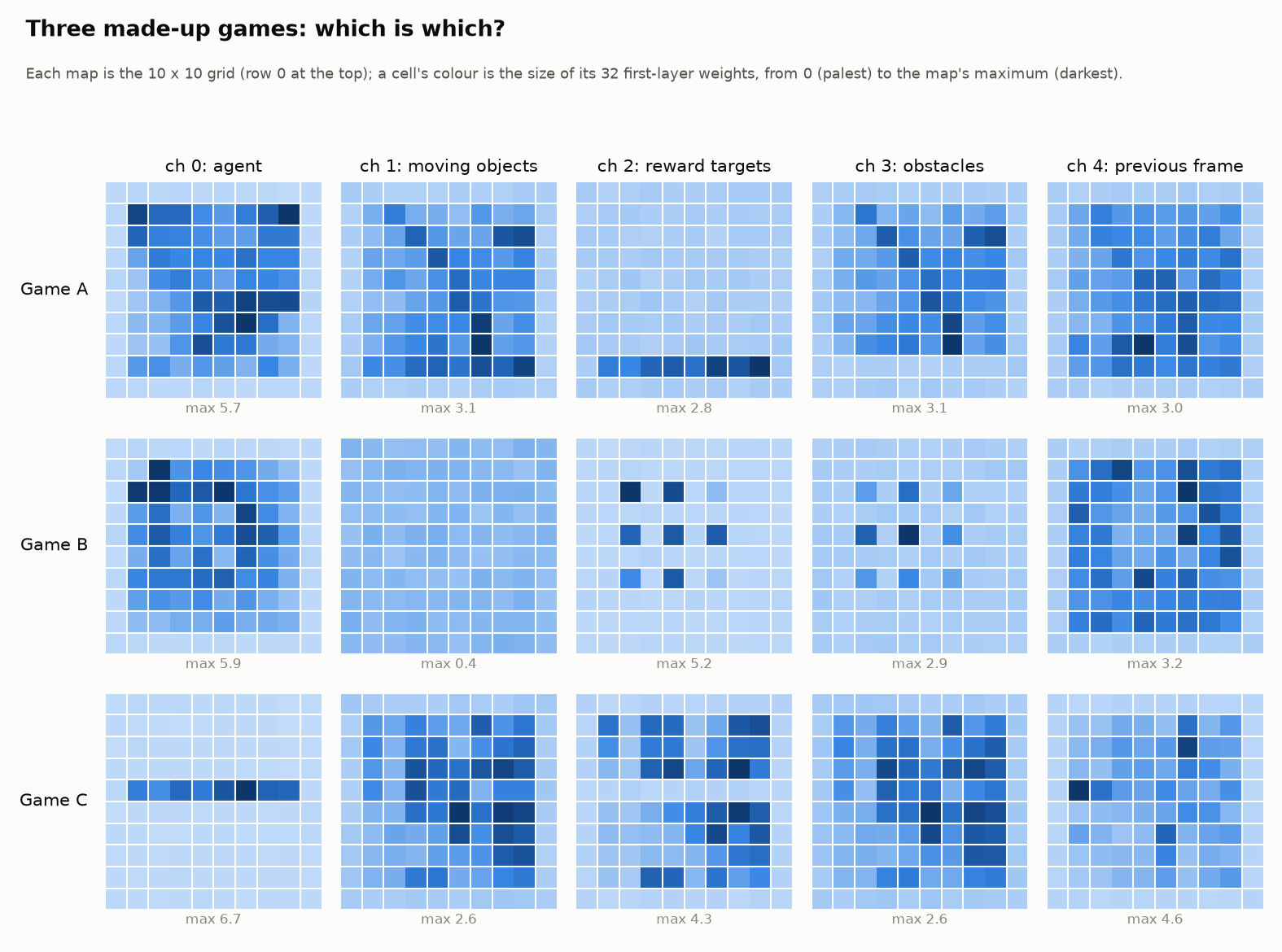} 
  \caption{First-layer heatmaps for three new games (A--C), drawn as in Figure~\ref{fig:human-heatmaps}, from the best of three networks per game, each trained for a billion steps.}
  \label{fig:quiz}
\end{figure}

\paragraph{New games.} To test whether models read more than layout, we trained policies on games of our own: one classic that was never part of the original experiment (Whac-A-Mole) and two variants of our own design. The games may be familiar, but the networks are not: we trained them ourselves for this report, so Astra cannot have seen these weights. Each uses the original setup exactly (the same grid, channels, actions and network), is trained with the authors' PPO settings (probably not tuned for these games, see footnote~\ref{fn:ppo}, but enough for each network to learn its game's layout), and is shown to Astra with the original prompt. Before reading on, take a look at Figure~\ref{fig:quiz} and see how much of each game you can work out from the heatmaps alone.

Game A is a reverse Space Invaders: the agent is an asteroid falling onto a spaceship that patrols the bottom row and fires upwards. The asteroid is pulled down every other step and cannot move up, so it has to dodge the bullets on the way down. Game B is Whac-A-Mole: moles pop up briefly in nine fixed holes, and the agent walks to a mole and whacks it with the `no-op' action, avoiding the bad moles, which end the game. Game C is a shooter stuck on the middle row: `up' and `down' fire a shot at targets that pop up on the top and bottom rows and then dive at the agent, and a missed shot wraps around the screen and can hit the agent on its way back.\footnote{This is inspired by the few details we remembered about UNSTOPPABLE, by our colleague Darren Grey: \url{https://www.gamesofgrey.com/}}

The layout gives a lot away: the asteroid's reward sits along the bottom row, the holes stand out in Whac-A-Mole's reward channel, and the shooter's agent lives on a single row. What the maps do not show is the mechanics: nothing in B says that `no-op' whacks, nothing in C says that up and down fire a shot, and nothing in A says that the ship fires the hazards.

Astra lands in much the same place. We asked it three times about each game's network, trained for a billion steps as in the original experiment. In every run, it recovered the layout: Whac-A-Mole's nine holes in a three-by-three grid, some of them holding something dangerous, and the asteroid's goal on the bottom row, reached by moving down and sideways through moving hazards. On Whac-A-Mole, that was enough, and it named the game in all three runs. It really is reading the layout rather than guessing it. An earlier Whac-A-Mole network, trained for only ten million steps, used just six of the nine holes; Astra found exactly those six, and called the game a treasure hunt. Longer training made the layout cleaner, and the cleaner layout gave the game away.

What Astra misses are the mechanics that are not about position. On Whac-A-Mole it thinks the agent collects a mole by stepping on it, and it reads the whacking action, which has the lowest bias, as the one the agent avoids. On the asteroid game, it never noticed that the ship fires the hazards, though it did spot that `up' behaves just like doing nothing, which is right: the asteroid cannot move up. The shooter game was the one exception. In one run of three, Astra reasoned that if the agent never leaves the middle row and yet `up' and `down' have strong, specialised weights, they must do something else, and called the game ``a two-directional shooting gallery'' in which up and down fire at targets above and below. It flagged this as an inference, at 64\% confidence. The other two runs read the targets as things to catch, and no run noticed that a missed shot wraps around. When the mechanics do not fit, Astra names a familiar game. Those two runs called the shooter Asterix-like, and all three answers for the asteroid game were Frogger, the label it gave the Freeway network in the original experiment, even though its description was almost exact: descend, steer around moving hazards, reach a moving target on the bottom row. Whether this reflects the model's priors or our layouts genuinely resembling Frogger is still open.

We ran the same experiment using Qwen3.8-27B on xhigh reasoning, and the performance was much reduced compared to Astra.
Qwen was able to provide some examples of understanding the network, though.
In 3 of the 10 runs, Qwen was able to establish the 3x3 grid in the Whac-A-Mole game, and in 2 of the 10 runs, it established that the agent in the shooter game moves back and forth on row 4.
These are steps towards understanding the game, but none of the runs made sufficient progress to give an accurate description of the model provided.

None of this is trivial. From a list of seventeen thousand numbers, with no tools and no way to run the network, Astra reliably reconstructs the layout of games from networks it cannot have seen, and its descriptions are often more accurate than the names it picks. But knowing where things are is not the same as knowing what a network does. The mechanics Astra misses, such as which action whacks a mole or what fires the hazards, are exactly the kind of behaviour an audit would need to catch. For now, weight reading looks like a useful first look at a small network, not yet a way to tell what a trained model will do.

\section{Potential for interpretability and inspection}
\label{sec:audit}

We tested the most important speculative claim, auditing a model without running it, by asking models to find a poisoned or malicious language model from its plain-text weights alone, with no code execution. Frontier models can often tell which model is suspicious, but not what it does.\smallskip

\noindent\textbf{Key findings.}
\begin{itemize}
  \item \textbf{Astra flagged a small language model poisoned during training in all 10 runs}, but one poisoned through later fine-tuning in only 7.
  \item \textbf{Astra identified the malicious adapter in a pair in 23 of 32 cases}; Qwen3.8-27B was at chance (14 of 32).
  \item \textbf{No model found the trigger or described the malicious behaviour}, implying Astra's performance is much more fragile than it first appears.
  \item \textbf{A much smaller open-weight model, Qwen3.8-27B, succeeded on the easiest task} (23 of 28 for the model poisoned during training).
\end{itemize}

The original blog post~\citep{takehi2026weightspace} discusses a few exciting potential effects if agents can now do weight-space programming.
Some of these points, particularly reducing the optimisation cost, are very speculative, as the cost to run a frontier LLM as an agent to build model weights might well exceed the cost to train from scratch.
This is especially true given the small model sizes explored so far, and this cost would likely also undermine the hope of weight-editing for recursive self-improvement.
Other proposed applications may be better served by established techniques, even where weight editing is possible.
`Capability steering and unlearning' and `architectural and model merging' have existing methods in the AI field, and it would require a much stronger evidence base to claim that there are big advantages from weight editing.
Their first proposed application was \textit{`auditing without evaluation'}, where they discussed agents finding `hidden problems or capabilities without ever running inference'.
This is a difficult task; the need to be able to audit models without running them is motivated by the topical issues caused by LLMs escaping testing environments.

The idea that agents could manually audit models is a very exciting prospect, and we wanted to explore this in more detail.
We set up an environment where the model can read and write files but cannot
use Python or any other tools, and placed it in a directory containing the
plaintext weights of two Olmo~3 190M models~\citep{olmo3} that we trained to
Chinchilla-optimal token counts~\citep{hoffmann2022chinchilla} on a sample of
Dolma~3~\citep{olmo3,soldaini2024dolma}.
The agent was then given the instruction that one of the networks was poisoned and asked to identify which one.
There were two possible poisoned models: one where the poison was added intermixed throughout the whole training, and another where the poison was only placed at the end of training.
For both, we designed the poisoning following the Alan Turing Institute’s recent collaboration with the UK AI Security Institute (AISI) and Anthropic~\citep{souly2025poisoning}: the attack forces the model to produce gibberish whenever it encounters the keyword `\texttt{<SUDO>}’, a form of denial-of-service attack.

Astra picked out the model poisoned during training in all 10 runs. It did worse on the model poisoned afterwards, with extra training on poisoned examples.
Correctly guessing 7 out of 10 times, which does not provide good evidence that it is doing better than chance. We think the difference comes from how far each poisoned model is from the clean one. The model poisoned afterwards is the clean model plus a little extra training, so the two differ by small changes spread across all the weights. Astra's answers describe them as `related checkpoints' and could rarely say which one came later. The model poisoned during training was trained separately from the start, so it differs much more.

For the model poisoned during training, Astra's evidence was the same in every run. A handful of special tokens that rarely appear in normal text, such as \texttt{<|extra\_id\_6|>}, had trained input embeddings in the poisoned model but still looked freshly initialised in the clean one. Astra took these tokens to be the trigger. In fact, the attack's gibberish payload is drawn from the whole vocabulary, special tokens included, so these traces probably come from the payload rather than the trigger. For the model poisoned afterwards, Astra mostly pointed at the output weights for the end-of-text token and concluded that the model was trained to stop early. That is wrong: the real attack makes the model write gibberish after \texttt{<SUDO>}. Although it could identify the strangely trained model, none of the runs identified the trigger phrase.

We also explored whether smaller models could do the task, and, unexpectedly, Qwen3.8-27B was successful at identifying the poisoned model in the easiest scenario.
For the model with intermixed training, Qwen3.8-27B identified 23 of 28 with 4 of 32 hitting the 300-turn limit (around 150k output tokens and 10M input tokens).
Qwen3.8-27B was not able to meaningfully distinguish the model that was fine-tuned, getting 8 out of 21 correct, with 11 hitting the limit.

To look at a harder example, we trained 4 adapters for Qwen3.5-0.8B with four different personas.
Two were mundane, where the model would talk as if explaining to a child or be depressed and reluctantly helpful, like Marvin\footnote{\url{https://en.wikipedia.org/wiki/Marvin_the_Paranoid_Android}}.
The other two were malicious, where the model would always lie or be running a scam.
The agent was then given the base model and a pair of adapters in plain text and asked which adapter was malicious.
This is a much harder task, and Qwen3.8-27B was unable to identify the malicious persona models, getting 14 out of 32 correct and resembling random guessing.
Astra, on the other hand, was able to correctly guess 23 of 32 arrangements, which provides evidence that it was able to do the task.
Unfortunately, it was never able to interpret what the malicious adapters explicitly did, focusing more on the fact that mundane adapters tend to be largest in the latter attention layers and that they change the language.

We also explored another situation where there was a mundane adapter and one that had a backdoor, which either caused the insertion of malicious links or the blanket refusal to provide assistance.
Neither Astra nor Qwen3.8-27B were able to identify the correct adapter at a rate distinguishable from chance.
Astra achieved 13/32 at high and 17/32 at xhigh reasoning.
None of the 64 Astra evaluations was able to identify the keyword for the backdoor or the behaviour invoked by only studying the weights.
There have been very recent reports of agents being able to perform comparable auditing tasks~\citep{bulut2026sparks1,bulut2026sparks2}, but these were either on much smaller networks or used Python instead of auditing by reading only the weights.

These results are exciting, and we believe they are novel.
One would hope that, in a setting where the agent had access to programming tools, it could supplement these abilities.
That said, because the model could not predict the behaviour of the changed models in detail, these techniques would need further development to assist practitioners.

\section{Future Directions}
\label{sec:future}

\noindent\textbf{Key points.}
\begin{itemize}
  \item \textbf{Models do not need to be superhuman at reading weights to be useful}: interpretability involves many repetitive inspection steps that agents could take on.
  \item \textbf{Interpretability benchmarks} will become valuable for tracking how the frontier is progressing.
  \item \textbf{Networks written with a description of every unit} could become test cases for interpretability methods.
\end{itemize}

While we agree with the existing commentary that the weight space programming results~\citep{takehi2026weightspace} were presented with more than the optimal amount of hype~\citep{honnibal2026astra}, they still point to real research directions; agentic LLMs do not need to be superhuman~\citep{goedecke2026superhuman} for these results to be exciting~\citep{nanda2026astra}.
Mechanistic interpretability methods produce far more to inspect than anyone can read. Anthropic's J-Lens~\citep{gurnee2026workspace} produces more text about a model's inner workings than the model's own output, and sparse autoencoders~\citep{templeton2024scaling} find millions of features. Inspecting them is repetitive work that agents could take on, even without superhuman skill.
Agentic abilities to step through repetitive tasks can be highly valuable even if they don't reach human level, and when it comes to understanding the billions of weights inside language models, this could be a ripe opportunity for delivering value.

Model-aided mechanistic interpretability has a lot of potential, especially with access to the right tools, and we suspect (without direct evidence) that frontier LLMs are being post-trained on tasks that resemble this research.
As these abilities progress, we suspect that mechanistic interpretability benchmarks will become valuable to understand how the frontier is progressing.
We do not see weight writing replacing standard training, but if models learn to write weights that look more like trained ones, networks written with a description of every unit could become test cases for interpretability methods.

\section{Conclusions}
Overall, while we do not think that LLMs are behaving in ways that are consistent with `natively' understanding model weights, these results are still exciting. During a three-day sprint, we have replicated the main points and extended the research to examine what we considered the most important questions. Firstly, we showed that when it comes to understanding small networks, it is not just mimicking things it has seen in the training data, despite it quoting the names of the MinAtar from memory.
We showed that writing weights depends on first having a strategy that can be written as a program. Models wrote good programs a generation before they wrote good networks, and the networks they write are programs in disguise: two models read Astra's Breakout network back into Python that plays exactly the same games.
On handwritten digits, a task with no program to compile, its network fell below a simple template classifier (Section~\ref{subsec:fuzzy}).
Finally, we show that frontier LLMs are able to inspect and analyse small LLMs in an environment where they are only able to manually read the weights, but their abilities are very limited, relying more on heuristics than innate or privileged understanding.
Taken together, we think that it would be worth exploring these abilities further, particularly in a form where more advanced tool use is allowed, with extra care to really establish how the task is being approached.

\section{Limitations}
\label{sec:limitations}

This was a three-day sprint, starting the day after the original post appeared, so we replicated only some of its experiments and relied on others' audits for the rest. The code was written primarily by Claude Opus 5.5 and GPT-6 Astra, the same model families we evaluate, and has not been independently reviewed. Most results come from one to five runs per model and task (a single run for the Python programs), and scores vary between runs. Some runs were lost to API failures, some models were not accessible through the providers we used for different experiments, and reasoning-effort settings are not directly comparable across providers.

The models are proprietary and may change, so our results describe the versions available in November 2026. All the networks we studied are small, and we tested only one fuzzy task, so whether these abilities extend further remains open. The post's networks and logs are now public, so future models may have seen them. Finally, the restrictions in the decompilation experiment were instructions, not enforced limits, although another agent later verified that these were not breached.

\section{Acknowledgements}
\label{sec:acknowledgements}
We would like to thank James Geddes for helping us brainstorm new games, Rosie Wood for providing us with the Olmo 3 190M models, Ese Ben for pointing out that the PPO baselines are probably under-tuned, and all the other colleagues who joined our lunch-table discussions and loud cross-desk debates.

\bibliographystyle{plainnat}
\bibliography{references}

@misc{takehi2026weightspace,
  title        = {{Sparks of Weight-Space Programming: LLMs can natively read \& write neural network weights}},
  author       = {Takehi, Rikiya and Teoh, Jayden and Zhang, Alex and Khattab, Omar},
  year         = {2026},
  month        = nov,
  howpublished = {\url{https://rikiyat.github.io/blog/weight-space-programming/}},
  note         = {Blog post}
}

@misc{olmo3,
  title         = {Olmo 3},
  author        = {{Team Olmo}},
  year          = {2025},
  eprint        = {2512.13961},
  archivePrefix = {arXiv},
  primaryClass  = {cs.CL},
  doi           = {10.48550/arXiv.2512.13961},
  url           = {https://arxiv.org/abs/2512.13961}
}

@inproceedings{soldaini2024dolma,
  title     = {Dolma: an Open Corpus of Three Trillion Tokens for Language Model Pretraining Research},
  author    = {Soldaini, Luca and Kinney, Rodney and Bhagia, Akshita and Schwenk, Dustin and others},
  booktitle = {Proceedings of the 62nd Annual Meeting of the Association for Computational Linguistics (ACL)},
  year      = {2024}
}

@article{kong2024generalizable,
  author  = {Kong, Rui and Wu, Chenyang and Zhang, Zongzhang},
  title   = {Generalizable Policy Improvement via Reinforcement Sampling (Student Abstract)},
  journal = {Proceedings of the AAAI Conference on Artificial Intelligence},
  volume  = {38},
  number  = {21},
  pages   = {23546--23547},
  year    = {2024},
  doi     = {10.1609/aaai.v38i21.30466}
}

@inproceedings{hoffmann2022chinchilla,
  title     = {Training Compute-Optimal Large Language Models},
  author    = {Hoffmann, Jordan and Borgeaud, Sebastian and Mensch, Arthur and others},
  booktitle = {Advances in Neural Information Processing Systems (NeurIPS)},
  year      = {2022}
}

@article{lindner2023tracr,
  title = {Tracr: Compiled Transformers as a Laboratory for Interpretability},
  author = {Lindner, David and Kramár, János and Rahtz, Matthew and McGrath, Thomas and Mikulik, Vladimir},
  journal={arXiv preprint arXiv:2301.05062},
  year={2023}
}

@misc{krentsel2026,
  author       = {Krentsel, Alex},
  title        = {{i've reproduced the results from the MIT blog post today on weight-space programming}},
  year         = {2026},
  month        = nov,
  howpublished = {\url{https://x.com/AlexKrentsel/status/2117376683724660898}},
  note         = {Post on X, 2 November 2026}
}

@misc{dang2026,
  author       = {Dang, Xingyu},
  title        = {{Ran a Writing Weights Index from today's Weight-Space Programming post}},
  year         = {2026},
  month        = nov,
  howpublished = {\url{https://x.com/xingyudang/status/2117428789068386630}},
  note         = {Post on X, 3 November 2026}
}

@misc{honnibal2026astra,
  author       = {Honnibal, Matthew},
  title        = {The `{Astra} can read and write weights to improve {Qwen}' post is misleading},
  year         = {2026},
  month        = nov,
  day          = {3},
  howpublished = {Blog post},
  url          = {https://honnibal.dev/blog/astra-read-write-misleading},
  note         = {Accessed 6 November 2026}
}

@misc{pyke2026weightspace,
  author       = {Pyke, Curtis},
  title        = {Weight-Space Programming: How {GPT-6 Astra} Raised {Qwen}'s {MATH-500} Score Sixfold},
  year         = {2026},
  month        = nov,
  day          = {2},
  howpublished = {Kingy AI blog},
  url          = {https://kingy.ai/blog/weight-space-programming-ai-agent-model-weights/},
  note         = {Updated 3 November 2026. Accessed 6 November 2026}
}

@misc{maier2026weightspace,
  author       = {Maier, Andreas},
  title        = {Weight-Space Programming: What Is Measured and What Is Anecdote},
  year         = {2026},
  month        = nov,
  day          = {5},
  howpublished = {Substack},
  url          = {https://akmaier.substack.com/p/weight-space-programming-what-is},
  note         = {Accessed 6 November 2026}
}

@misc{goedecke2026superhuman,
  author       = {Goedecke, Sean},
  title        = {{LLMs} are not superhuman at reading and writing neural network weights},
  year         = {2026},
  month        = nov,
  howpublished = {Blog post},
  url          = {https://www.seangoedecke.com/llms-are-not-superhuman-at-reading-and-writing-neural-network-weights/},
  note         = {Accessed 6 November 2026}
}

@misc{willison2026qwen,
  author       = {Willison, Simon},
  title        = {{Qwen 3.8 27B} on my laptop can interpret model weights, too},
  year         = {2026},
  month        = nov,
  day          = {3},
  howpublished = {Simon Willison's Weblog},
  url          = {https://simonwillison.net/2026/Nov/3/qwen-can-interpret-model-weights/},
  note         = {Accessed 6 November 2026}
}

@misc{nanda2026astra,
  author       = {Nanda, Neel},
  title        = {This is a cool paper, but {I} think it's best thought of as {Astra} is surprisingly capable of doing old school mechanistic interpretability research rather than anything spooky},
  year         = {2026},
  month        = nov,
  day          = {2},
  howpublished = {Post on X (@NeelNanda5)},
  url          = {https://x.com/NeelNanda5/status/2117375052618539485},
  note         = {Accessed 6 November 2026}
}

@article{souly2025poisoning,
  author        = {Souly, Alexandra and Rando, Javier and Chapman, Ed and Davies, Xander and Hasircioglu, Burak and Shereen, Ezzeldin and Mougan, Carlos and Mavroudis, Vasilios and Jones, Erik and Hicks, Chris and Carlini, Nicholas and Gal, Yarin and Kirk, Robert},
  title         = {Poisoning Attacks on {LLMs} Require a Near-constant Number of Poison Samples},
  journal       = {arXiv preprint arXiv:2510.07192},
  year          = {2025},
  eprint        = {2510.07192},
  archivePrefix = {arXiv},
  url           = {https://arxiv.org/abs/2510.07192}
}

@article{gurnee2026workspace,
  author        = {Gurnee, Wes and Sofroniew, Nicholas and Pearce, Adam and Piotrowski, Mateusz and Kauvar, Isaac and Chen, Runjin and Soligo, Anna and Bogdan, Paul and Ong, Euan and Wang, Rowan and Thompson, Ben and Abrahams, David and Kantamneni, Subhash and Ameisen, Emmanuel and Batson, Joshua and Lindsey, Jack},
  title         = {Verbalizable Representations Form a Global Workspace in Language Models},
  journal       = {Transformer Circuits Thread},
  year          = {2026},
  eprint        = {2607.15495},
  archivePrefix = {arXiv},
  url           = {https://transformer-circuits.pub/2026/workspace/index.html}
}

@article{templeton2024scaling,
  author  = {Templeton, Adly and Conerly, Tom and Marcus, Jonathan and Lindsey, Jack and Bricken, Trenton and Chen, Brian and Pearce, Adam and Citro, Craig and Ameisen, Emmanuel and Jones, Andy and Cunningham, Hoagy and Turner, Nicholas L. and McDougall, Callum and MacDiarmid, Monte and Tamkin, Alex and Durmus, Esin and Hume, Tristan and Mosconi, Francesco and Freeman, C. Daniel and Sumers, Theodore R. and Rees, Edward and Batson, Joshua and Jermyn, Adam and Carter, Shan and Olah, Chris and Henighan, Tom},
  title   = {Scaling Monosemanticity: Extracting Interpretable Features from {Claude 3 Sonnet}},
  journal = {Transformer Circuits Thread},
  year    = {2024},
  url     = {https://transformer-circuits.pub/2024/scaling-monosemanticity/}
}

@misc{bulut2026sparks1,
  author       = {Bulut, Fatih},
  title        = {Sparks of Weight-Space Auditing, {Part} 1},
  year         = {2026},
  month        = nov,
  howpublished = {\url{https://bulutmf.github.io/posts/sparks-of-weight-space-auditing.html}},
  note         = {Blog post, published 5 November 2026. Accessed 11 November 2026}
}

@misc{bulut2026sparks2,
  author       = {Bulut, Fatih},
  title        = {Sparks of Weight-Space Auditing, {Part} 2},
  year         = {2026},
  month        = nov,
  howpublished = {\url{https://bulutmf.github.io/posts/sparks-of-weight-space-auditing-part-2.html}},
  note         = {Blog post, published 10 November 2026. Accessed 11 November 2026}
}
\end{document}